\documentclass[runningheads]{llncs}

\usepackage{eccv}

\usepackage{eccvabbrv}
\usepackage{subcaption}
\usepackage{graphicx}
\usepackage{booktabs}
\usepackage{wrapfig}
\usepackage[accsupp]{axessibility}  
\usepackage{booktabs}
\usepackage[table]{xcolor}

\usepackage{hyperref}
\usepackage{caption}
\usepackage{booktabs}
\usepackage{svg}
\usepackage{graphicx}
\usepackage{color}
\usepackage{placeins}
\usepackage{orcidlink}

\begin{document}

\title{RefDiT: Local Attribute Guidance in Reference-Based Image Generation} 
\titlerunning{RefDiT: Local Attribute Guidance in Reference-Based Image Generation}

\author{Rameshwar Mishra\inst{1,}\thanks{Work done during internship at Adobe Research}\orcidlink{0009-0009-5368-3371} \and
Srikrishna Karanam\inst{2}\orcidlink{0000-0002-7627-7765} \and
A V Subramanyam\inst{1}\orcidlink{0000-0002-8873-4644}}

\authorrunning{R.~Mishra et al.}

\institute{Indraprastha Institute of Information Technology Delhi, India \\
\email{\{rameshwarm, subramanyam\}@iiitd.ac.in}
\and
Adobe Research\\
\email{skaranam@adobe.com}}

\maketitle
\begin{figure}
    \includegraphics[width=\textwidth]{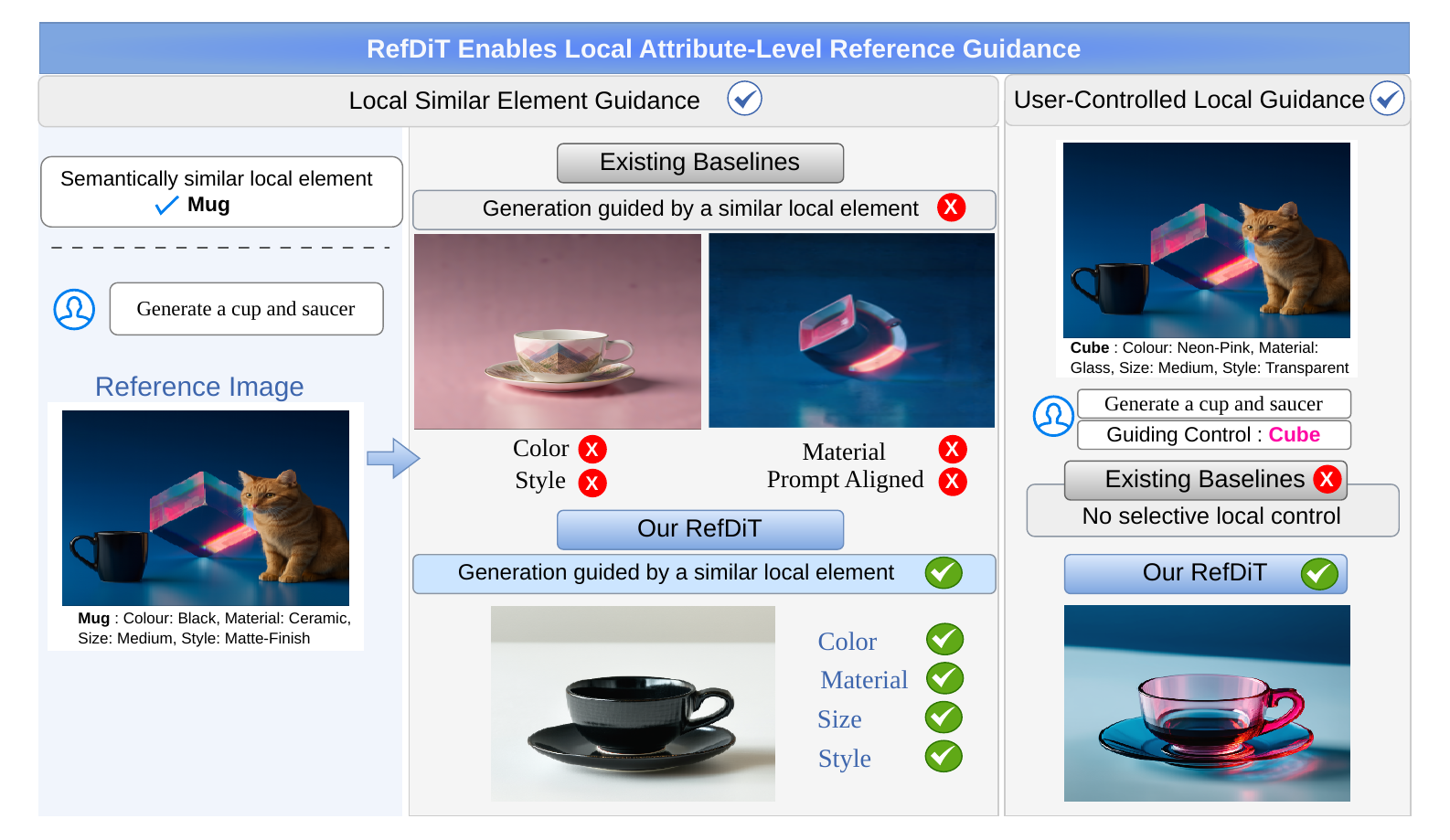}
    \caption{\textbf{RefDiT} guides reference-based image generation using attributes of semantically matched local elements. Unlike global conditioning approaches, our triplet-conditioned LoRA training for Diffusion Transformer (DiT) blocks establishes region-level correspondence between attribute tokens and their visual regions, enabling selective local attribute guidance.}
    \vspace{-35pt}
    \label{ref:Teaser}
\end{figure}


\begin{abstract}
Personalization models generate images guided by a few subject references, while style transfer methods aim to produce images aligned with a global style derived from a reference image. Recent approaches perform well when the reference image contains a single object, effectively capturing a global style that encompasses all implicit attributes. However, when applied to complex real-world scenes containing multiple objects with distinct attributes, existing methods struggle to selectively control generation using attributes from specific local regions of the reference image. Due to their primarily global conditioning strategies, these methods cannot reliably guide generation using attributes from specific local elements in the reference image. Moreover, existing methods typically employ a single identifier token to capture all details from the reference, resulting in a lack of individual, attribute-level control.

Motivated by these limitations, we propose \textbf{RefDiT}, a novel framework for reference-guided image generation. RefDiT takes as input a reference image, a text prompt, and an optional user-provided guidance control. RefDiT employs local region guidance using the attributes of local elements. It constructs an attribute-aware conditioning signal from the reference image by performing attribute-level decomposition of the identifier tokens to train low-rank adapter (LoRA) blocks of a diffusion transformer (DiT)-based generative model. RefDiT leverages multi-modal joint attention to learn the correspondence between identifier tokens and local regions in the reference image, enabling more effective local guidance. RefDiT achieves a local attribute matching score of 0.88, outperforming state-of-the-art methods such as UnZipLoRA (0.54), K-LoRA (0.58), and B-LoRA (0.42). Compared to commercial models like Gemini-Banana (0.74) and GPT-5 (0.84), user studies indicate that participants find RefDiT’s outputs to be of similar quality.
\end{abstract}
\section{Introduction}

\label{sec:intro}
Diffusion models \cite{ho2020denoising,rombach2022high} have demonstrated strong capabilities in generating diverse and high-quality images. While text has been the most common form of guidance for controlling the generation process, the limitations of natural language in describing nuanced scenes have motivated the development of reference-guided generation. DreamBooth \cite{ruiz2023dreambooth} and StyleDrop \cite{sohn2023styledrop} train diffusion models to capture both content and style details from subject images. B-LoRA \cite{frenkel2024implicit}, K-LoRA \cite{ouyang2025k}, and ZipLoRA \cite{shah2024ziplora} introduce strategies to learn the global style and content by training a few low-rank adapter (LoRA) \cite{hu2022lora} blocks. Once trained, these weights can generate images that combine style and content from different references. UnZipLoRA \cite{liu2024unziplora} decomposes a single image into content and style concepts, enabling effective text-conditioned generation guided individually by either the content or the style. ProSpect \cite{zhang2023prospect} and MATTE \cite{agarwal2025image} inject attribute-level details from a reference into different stages of the generation.

Recent DiT-based editing methods, such as Flux.1 Kontext \cite{labs2025flux} and FlowEdit \cite{kulikov2024flowedit}, demonstrate strong performance in editing local properties. These models can modify attributes within local regions, but they are not well-suited for generating new components guided by localized regions in a reference image. One possible adaptation is to treat our task as a sequence of editing operations, starting from the reference image and applying a series of edits to arrive at the desired output. However, as reported in \cite{labs2025flux}, these methods lose a substantial amount of the original source information after only a few editing steps, often resulting in distorted images.

However, reference images contain far more information than just global style and content. Consider real-world images that include multiple objects, where a user may wish to generate a new image guided by attribute-level details of a specific local element in the reference image. In such cases, critical and substantial guidance can come from the attributes of local regions, particularly when generating novel images containing similar local elements.
\begin{wrapfigure}{r}{0.48\textwidth}
   \vspace{-25pt}
    \centering
    \includegraphics[width=\linewidth]{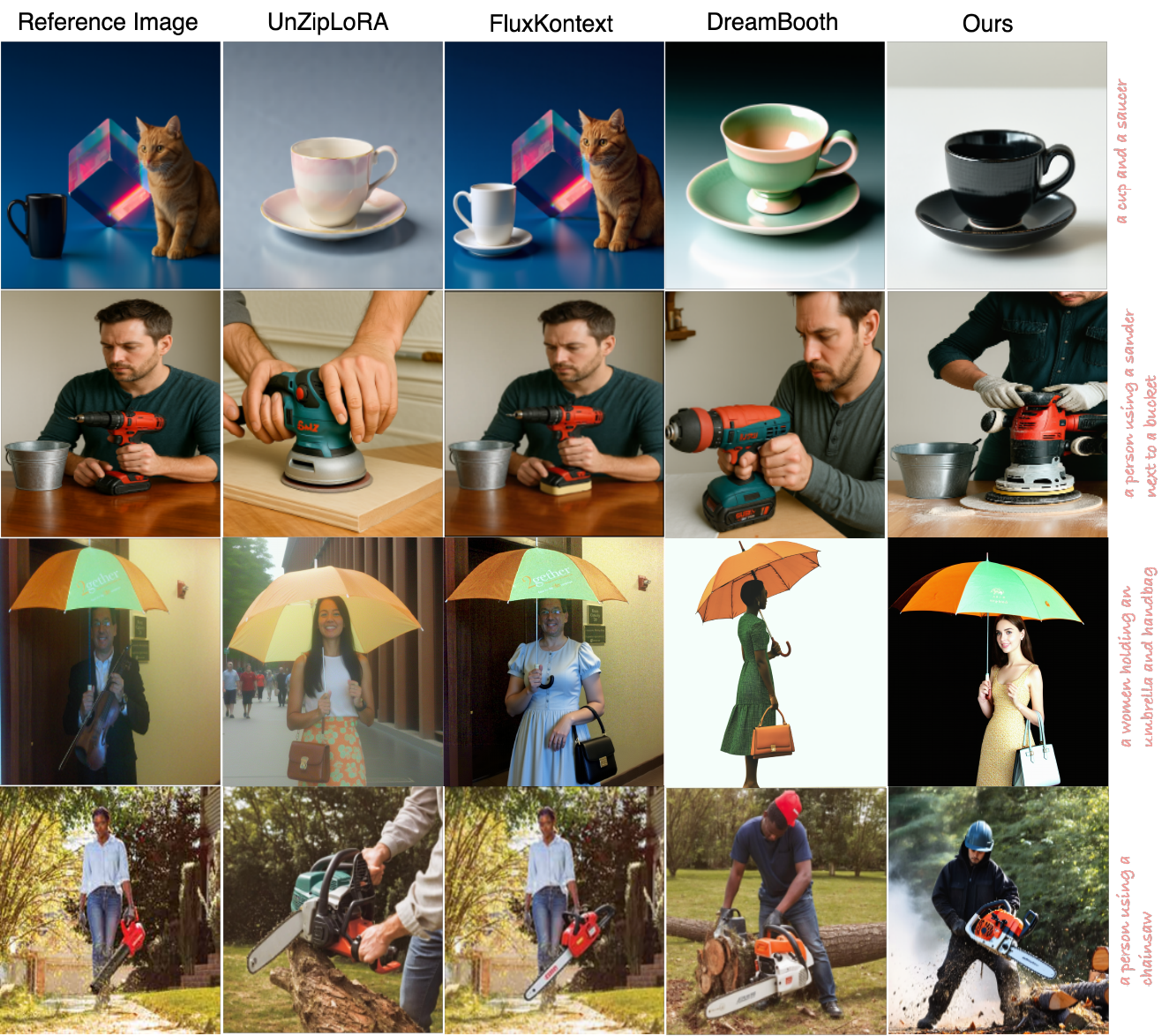}
    \caption{Limitations of global conditioning in existing reference-guided image generation methods.}
    \vspace{-25pt}
    \label{fig:analysis_1} 
\end{wrapfigure}
When we analyzed the performance of personalization methods and LoRA-based stylization approaches on complex real-world scenes involving interactions among multiple objects, we observed that these methods often fail to provide effective local guidance. Existing approaches rely on global-level image conditioning along with a single identifier to capture all details of the elements present in the image. This design results in a lack of both local and attribute-level control. 

As shown in Figure~\ref{fig:analysis_1}, row 1 presents a reference image containing three objects, a cube, a cat, and a coffee mug. When generating a new image with a prompt \textcolor{red}{``cup and saucer''}, existing methods fail to localize the mug, the most similar local element to a cup and saucer, and are unable to guide the generation based on local attributes such as the mug's color and material. Furthermore, there is no control for the user to specify which local elements and their corresponding attributes should guide the generation process. In contrast, Figure~\ref{ref:Teaser}, column 3, shows results from our method, where users can explicitly provide such controls, enabling more precise guidance.

Through extensive experiments on multi-object reference scenes, we identify a critical research gap in existing reference-based image generation methods. Current approaches struggle to generate images where local attribute concepts from the reference should guide the generation process. In this work, we propose \textbf{RefDiT}, a method in which relevant local regions from the reference image guide the generation process. For example, in the second row of Fig.~\ref{fig:analysis_1}, when the user wants to generate a \textcolor{red}{``person using a tool (sander)''}, RefDiT produces an image where attributes such as the color, material, appearance, and size of the new tool are guided by a similar tool (drill) found locally in the reference. RefDiT extracts structured object triplets from the reference image using a multi-modal large language model (MLLM), capturing both object-level attributes and inter-object relationships. This triplet formulation provides a semantic representation that does not require explicit spatial supervision, such as bounding boxes or segmentation masks. The extracted triplet-level information is then used to construct both global and local guidance signals for the generation model through attribute-aware identifier tokens. Furthermore, RefDiT introduces a mechanism to update the multi-modal DiT weights to learn correspondences between these identifier tokens and their associated local regions in the reference image.

We summarize our contributions as follows:
\begin{itemize}
\item We identify and show a critical gap in existing reference-based image generation methods, showing that they have limited capacity to provide effective attribute-aware, local-level guidance when the reference image contains multiple objects with distinct attribute properties.
\item We propose \textbf{RefDiT}, which approaches reference guidance differently from prior methods. RefDiT leverages triplet-level information from the reference image to curate simple yet effective guidance signals for generative models. We introduce a triplet-conditioned LoRA training for DiT-based generative models that establishes region-level correspondences between identifier tokens and their corresponding visual regions in the reference image.

\item We demonstrate the effectiveness of RefDiT across a diverse set of references. Through extensive qualitative and quantitative comparative analysis with LoRA-based personalization methods, DiT-based editing approaches, and commercial image generation models, we demonstrate that RefDiT yields consistent improvements in reference-guided image generation.
\end{itemize}

\section{Related Work}
\label{sec:related_work}

\textbf{Diffusion fine-tuning for customization.} 
With recent advancements in large-scale text-to-image (T2I) models, several methods have been developed to customize outputs based on user-provided references. Textual Inversion~\cite{gal2022image} captures reference details through text embeddings, while DreamBooth~\cite{ruiz2023dreambooth} fine-tunes the entire model to learn subject-specific details. Methods such as~\cite{han2023svdiff} and~\cite{kumari2023multi} learn visual concepts by optimizing specific parts of the diffusion model. ProSpect~\cite{zhang2023prospect} and MATTE~\cite{agarwal2025image} adapt textual inversion by injecting different conditioning signals at various stages of generation to provide attribute-level guidance.

\noindent \textbf{LoRA training for reference guidance.} 
LoRA~\cite{hu2022lora} weights and their variants~\cite{hayou2024lora+,ren2024melora,kopiczko2023vera} have recently demonstrated strong effectiveness across a variety of tasks. B-LoRA~\cite{frenkel2024implicit} analyzes different blocks of the SDXL U-Net architecture~\cite{podell2023sdxl} to define content and style components. It trains content and style LoRAs independently, allowing either to guide the generation process. K-LoRA~\cite{ouyang2025k} employs adaptive weight selection using pre-trained LoRAs for content and style references. ZipLoRA~\cite{shah2024ziplora} merges LoRA weights through fusion matrices and hyperparameter tuning, while UnZipLoRA~\cite{liu2024unziplora} uses prompt and block separation strategies to disentangle content and style characteristics, enabling independent guidance during generation.  
Although these methods are effective, they lack the ability to guide generation based on local regions of the reference. They condense all image properties into a global style representation, thus limiting the granularity of guidance. 

\noindent \textbf{DiT-based editing methods.} 
Advances in diffusion models have led to numerous image editing approaches~\cite{hertz2022prompt,meng2021sdedit,li2025instructany2pix,wang2025editclip}. Recently, DiT-based image editing models have shown strong performance in localizing edits. Flux.1 Kontext~\cite{labs2025flux}, FlowEdit~\cite{kulikov2024flowedit}, and~\cite{yin2025training} can effectively modify attributes of locally present elements in an image. However, these methods are not suitable for our task, where the goal is to guide the generation of a new image based on locally present elements, rather than manipulating an existing one. Comparative results shown later demonstrate the inapplicability of these models for our setting.  
Commercial models such as GPT-5~\cite{openai-gpt5} and Nano-Banana~\cite{google-nanobanana} allow users to generate images guided by a reference, though the exact handling of guidance in these models is not publicly known. We also include comparisons with these systems.

\section{Method}
\label{sec:Method}
\subsection{Preliminaries}
\noindent \textbf{Rectified-Flow Models.} 
Flux and Stable Diffusion 3 employ multi-modal DiTs trained to parameterize rectified-flow models. Flow-based methods~\cite{lipman2210flow} map samples $x_1$ from a noise distribution $p_1$ to data samples $x_0$ from the data distribution $p_0$, while rectified flows~\cite{liu2022flow} learn ODEs along straight paths between $p_0$ and $p_1$, defined as: $z_t = (1 - t)x_0 + t\epsilon,\space \epsilon \sim \mathcal{N}(0, 1)$.
Flux and SD3 are optimized with the following conditional flow matching loss:
\begin{equation}
\frac{1}{2} \, \mathbb{E}_{t \sim \mathcal{U}(t), \, \epsilon \sim \mathcal{N}(0,I)}
\Big[w_t \lambda'_t \, \| \epsilon_\Theta(z_t, t) - \epsilon \|^2 \Big],
\end{equation}
where $\lambda'_t$ denotes the signal-to-noise ratio and $w_t$ is a time-dependent weight. $\epsilon_\Theta(z_t, t)$ is parameterized by a multi-modal diffusion transformer.

\noindent \textbf{Multi-Modal DiT Layer.} 
Multi-modal diffusion transformers (DiTs) extend standard transformer layers with multi-modal attention (MMATTN) that jointly process image patches and text tokens. Each layer operates on image embeddings $x \in \mathbb{R}^{h \times w \times d}$ and prompt embeddings $p \in \mathbb{R}^{l \times d}$.  
In MMATTN layers, modality-specific queries, keys, and values are projected for both image and text streams, denoted as $\{Q_x, K_x, V_x\}$ and $\{Q_p, K_p, V_p\}$, and are combined into a unified attention operation:
\begin{equation}
o_x, o_p = \text{softmax}(q_{xp} k_{xp}^\top) v_{xp},
\label{eqn_dit}
\end{equation}
where $q_{xp}$, $k_{xp}$, and $v_{xp}$ represent concatenated projections across modalities.  
The resulting outputs are modulated via adaptive layer normalization~\cite{xu2019understanding} and integrated through residual connections, producing updated embeddings $x^{L+1}$ and $p^{L+1}$ for the next layer. This dual-stream structure allows effective alignment of visual and textual information.
\subsection{RefDiT}
\textbf{Problem Setup.} 
Given a reference image $I$, a text prompt $P$, and an optional text based control $M_c$, the goal is to generate a new image where the generation process is guided by relevant local elements present in $I$. The selection of relevant elements can be determined either through $M_c$ or by predefined category matching, where local elements in the reference are matched with those in the target based on the semantic similarity of their text labels. Next, we describe our proposed method, RefDiT, which first prepares an attribute-aware guidance signal from the reference image and employs a triplet-conditioned LoRA training strategy for DiT blocks.
\begin{figure*}[t]
    \centering
    \includegraphics[width=\textwidth]{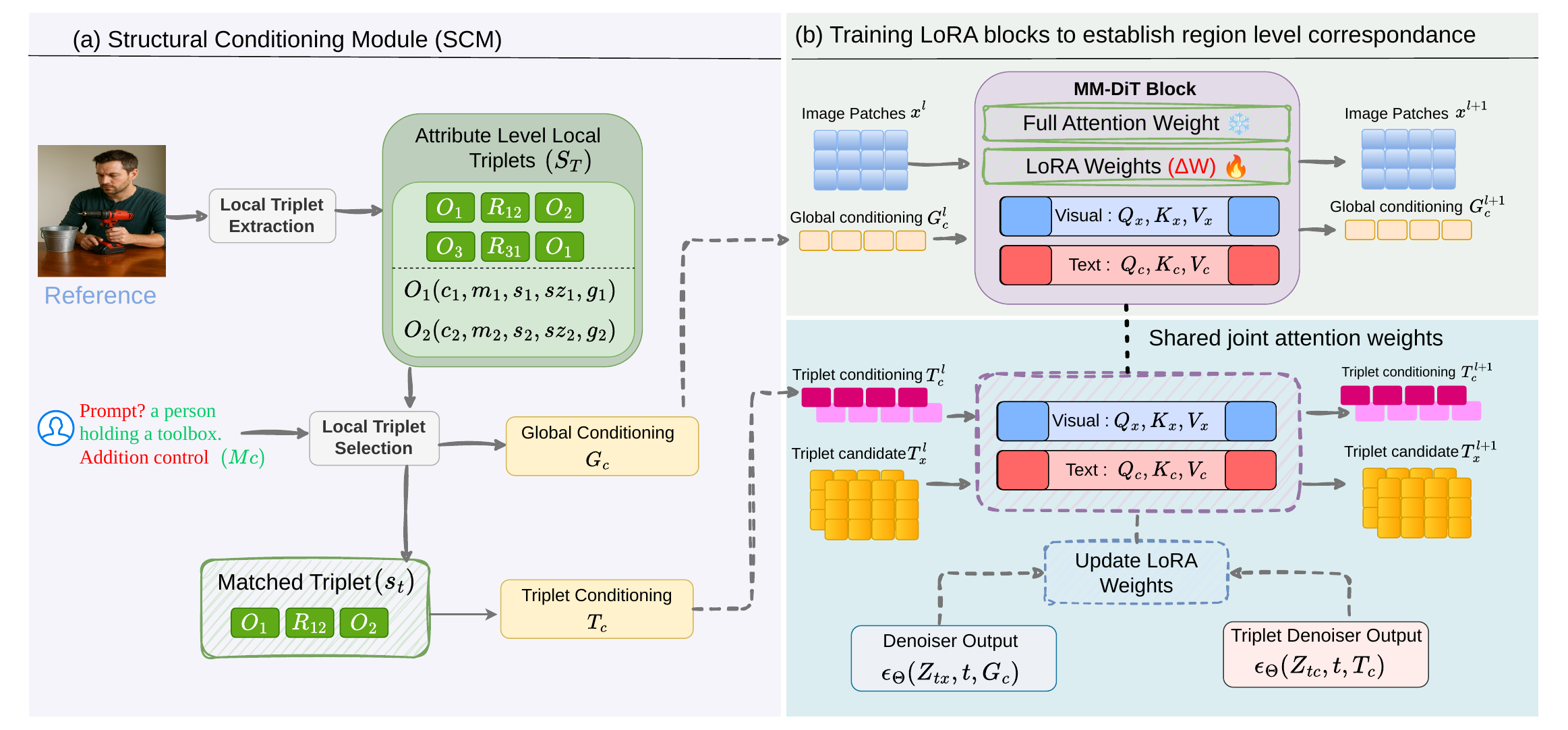}
    \caption{(a) SCM extracts attribute-aware object triplets from the reference image to construct global and local conditioning signals. (b) Triplet-conditioned LoRA training enables identifier tokens to learn region-level correspondence in DiT.}
    \vspace{-0.8cm}
    \label{fig:method}
\end{figure*}

\noindent \textbf{Structural Conditioning Module (SCM).}  
Existing personalization methods \cite{shah2024ziplora,ruiz2023dreambooth,liu2024unziplora}, in addition to the reference image, typically expect a brief textual description as input. Given this description, these methods introduce a single identifier token per image or per object to capture the details of the reference image, which may not be sufficient to represent nuanced, attribute-level details of real-world images containing multiple objects with diverse properties. Figure \ref{fig:analysis_1}, row 3, shows that UnZipLoRA generates the umbrella color based on the wall instead of the umbrella in the reference image, highlighting the ambiguity caused by the conditioning.
Frenkel \textit{et al.}~\cite{frenkel2024implicit} avoid using a description and simply create a conditioning text such as ``A [identifier token].''  
Given the limitations of existing strategies, we introduce a new approach that uses local triplets to generate more effective conditioning signals.

Figure~\ref{fig:method}(a) provides an overview of SCM.  
We use a multi-modal large language model (MLLM) based \textbf{local triplet extraction} to extract attribute-level triplets from the reference image. 
Given a reference image, we obtain a set of triplets $S_T$, where $n^{th}$ triplet $T_{c_n}$ is represented as $[O_i, R_{ij}, O_j]$, with $O_i$ and $O_j$ denoting the $i^{th}$ and $j^{th}$ objects in the image, and $R_{ij}$ representing the interaction between them.   
Each object is associated with a set of attributes, denoted for the $i^{th}$ object as $\{c_i, m_i, sz_i, sh_i, g_i\}$,  
corresponding respectively to the five key attribute types: $A_{\text{type}} = \{\text{color}, \text{material}, \text{size}, \text{shape}, \text{global appearance}\}$.  
In this work, similar to \cite{zhang2023prospect,agarwal2025image}, we limit ourselves to these five key attributes, and for a single image, we obtain $N$ such triplets. To capture intrinsic details not described by attributes such as color and material, we introduce a global appearance attribute, defined over a curated vocabulary of style descriptors (e.g., glossy, neon, grainy). This improves alignment with the appearance of the corresponding local element.  
Next, we create a global conditioning signal by introducing multiple attribute-level identifiers for objects in the image.  
We define an identifier as 
$``<attr\_\{type_j\}\_\{value^i_j\}>"$,  
where $type_j$ is the $j^{th}$ element of $A_{type}$ and $value^i_j$ is the value of the $j^{th}$ attribute for object $O_i$. We append these identifier tokens to each object and concatenate all triplets $(Tc_1, Tc_2, \dots, Tc_n)$ with their respective identifiers to generate a global conditioning signal $G_c$. Example of $G_c$ for a particular case is as follows:

\begin{figure*}[h]
    \centering
    \vspace{-18pt}
    \includegraphics[width=1\textwidth]{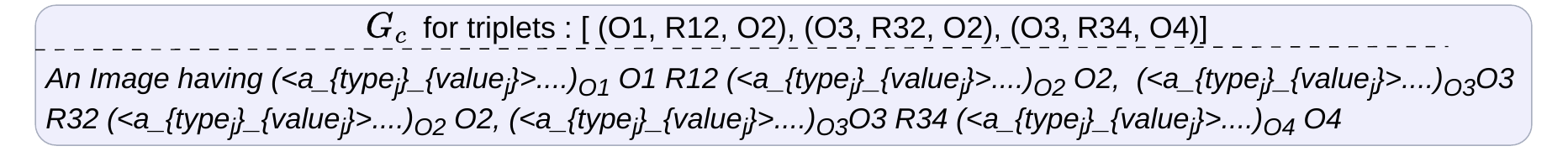}
\end{figure*}
\vspace{-24pt}
Using the set of triplets $(S_T)$ and an optional user-defined text based control $(M_c)$, we employ MLLM based \textbf{local triplet selection} to select a subset $s_t$ that contains local elements matching the inference prompt and relevant attributes constrained by the user provided control.  
If the control is not provided, we define it using ``category matching'', where triplet selection is guided by semantic similarity of object labels. We provide multiple rule based ambiguity resolving instructions to triplet selection MLLM to tackle ambiguous cases.    
We also include interactions among objects, as these contribute to improved guidance and control during the generation process by replicating interactions in the new images.
After obtaining matching triplets, we perform control adjustment in the inference prompt by adding attribute identifiers corresponding to the matched local elements from the reference image.

\noindent \textbf{Training LoRA Blocks to Establish Region-Level Correspondence.}

\begin{wrapfigure}{r}{0.5\textwidth} 
    \vspace{-25pt}
    \centering
    \includegraphics[width=\linewidth]{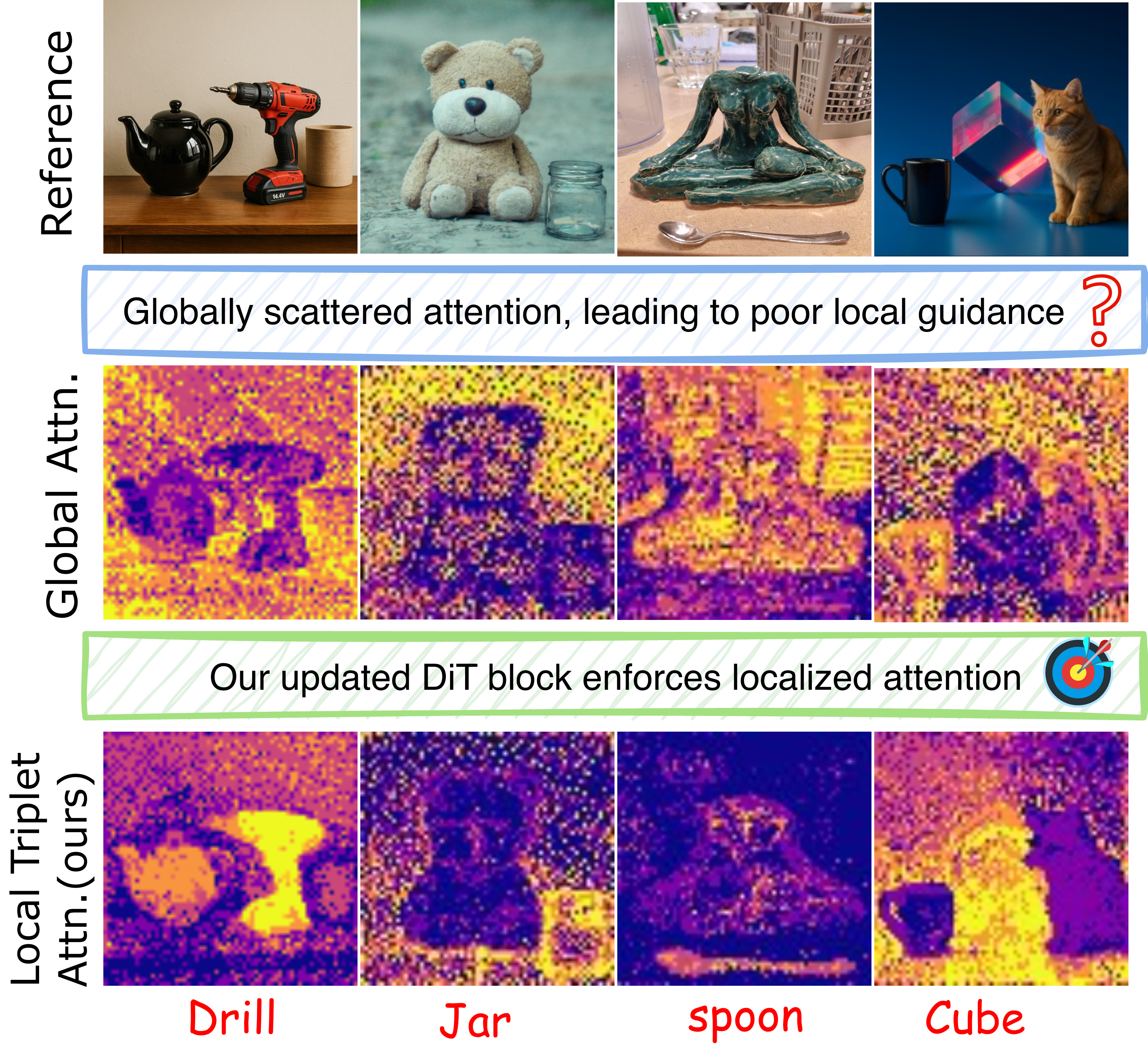}
    \caption{Comparison of attention maps illustrating local correspondence of identifier tokens for mentioned objects, where yellow denotes higher attention and black denotes lower attention.}
    \vspace{-25pt}
    \label{fig:fig_attention}
    
\end{wrapfigure}
After obtaining the triplet-level details from the reference and constructing the attribute-aware conditioning signal, the next step is to map the identifiers to their visual appearances in the reference image. 

We discover that simply training the weights with the new global conditioning causes identifier tokens to attend broadly across spatial regions, instead of focusing on their corresponding local objects. Figure \ref{fig:fig_attention} row 2 shows mean cross-attention map for identifier tokens in the global conditioning signal, it can be seen that they are not focusing on local objects effectively. To provide effective local guidance we utilize matched local triplets to train the LoRA blocks of our generation model. Figure \ref{fig:method}(b) illustrates how RefDiT integrates global and triplet-level conditioning using shared attention weights.

As described in the preliminaries, DiT jointly processes image and text modalities.  
Let $\{\mathbf{Q}_x, \mathbf{K}_x, \mathbf{V}_x\}$ denote the attention weights for the visual stream (patch features) and $\{\mathbf{Q}_c, \mathbf{K}_c, \mathbf{V}_c\}$ denote those for the text-conditioning stream at layer $l$.  
For our global conditioning feature $G_c$ and reference image patch features $x_r$, following Eq. \ref{eqn_dit}, the output of a DiT block is:
\begin{equation*}
    x^{l+1}_r, G^{l+1}_c
    =
    \text{softmax}\!\left(
    \mathbf{q}^l_{xc}
    {\mathbf{k}^l_{xc}}^\top
    \right)
    \mathbf{v}^l_{xc},
    \label{eqn:eqn_dit_method}
\end{equation*}
where $\mathbf{q}_{xc}, \mathbf{k}_{xc}, \mathbf{v}_{xc}$ are concatenated attention outputs from the individual streams, for example, 
\begin{equation*}
    \mathbf{q}_{xc}^l
    =
    \left[
    \mathbf{Q}_x \mathbf{x}_1^l,
    \;\dots,\;
    \mathbf{Q}_c \mathbf{c}_1^l,
    \;\dots
    \right]
\end{equation*} 
The DiT architecture fuses both modalities through a joint attention mechanism, where text features are collectively mapped to patch features in latent space.  
We utilize multi-modal fusion in DiTs to establish local region correspondence for identifier tokens in our text conditioning. From global-level features, we compute triplet region candidates as follows.  
Suppose SCM provides $K$ matching triplets, where the $i^{th}$ triplet is $Tc_i$.  
We define a corresponding triplet visual candidate for $Tc_i$ at layer $l$ as ${Tx_i}^l$, initialized with the reference patch features ${Tx}^0 = {x_r}^0$.  
We use the same attention weights $\{Q^l_x, K^l_x, V^l_x\}$ and $\{Q^l_c, K^l_c, V^l_c\}$ to update these triplet candidates in each layer as follows:
\begin{equation*}
    Tx_i^{l+1}, Tc_i^{l+1}
    =
    \operatorname{softmax}\!\left(
    \mathbf{q}_{txc}^l
    {\mathbf{k}_{txc}^l}^{\top}
    \right)
    \mathbf{v}_{txc}^l,
    \label{eqn:triplet_dit}
\end{equation*}
where $
\mathbf{q}_{txc}^l
=
\left[
\mathbf{Q}_x \, Tx_{i,1}^l,
\;\dots,\;
\mathbf{Q}_c \, Tc_{i,1}^l,
\;\dots
\right]
$, and $\mathbf{k}_{txc}^l, \mathbf{v}_{txc}^l$ are computed similarly. This modification to the DiT block allows a triplet candidate, initialized with reference patch features, to progressively guide attention toward corresponding local regions across layers through fusion with triplet-level text conditioning. Figure \ref{fig:fig_attention} row 3 shows cross attention maps of identifier tokens in the triplet conditioning after training. It can be seen that the identifier tokens attend local objects more dominantly. Importantly, triplet candidates are processed using shared DiT weights and serve as auxiliary attention streams during LoRA training, without modifying inference-time architecture. We utilize these triplet candidates to define our training objective, which enables efficient guidance of the generation process using relevant local regions.

\noindent \textbf{Training Objective.}  
We use a combination of two losses to train the LoRA blocks of our generation model.  
The first is a standard global reconstruction loss, which guides the weights to capture overall information from the reference image, defined as:
\begin{equation*}
    L_r = \frac{1}{2} \, \mathbb{E}_{t \sim \mathcal{U}(t), \, \epsilon \sim \mathcal{N}(0,I)}
\left[ w_t \lambda'_t \, \| \epsilon_\Theta(z_{tx}, t, G_c) - \epsilon \|^2 \right]
\end{equation*}
We also use the same network $\epsilon_\Theta$ with shared weights for triplet conditioning to denoise triplet candidates. To enable identifier tokens to attend to their corresponding local regions, we define the triplet region consistency loss as:
\begin{equation*}
    {L_c}_i = \frac{1}{2} \, \mathbb{E}_{t \sim \mathcal{U}(t), \, \epsilon \sim \mathcal{N}(0,I)}
    \left[ w_t \lambda'_t \, \| \epsilon_\Theta(z_{tc}, t, Tc_i) - \epsilon \|^2 \right]
\end{equation*}
If there are $K$ matching triplets, the overall triplet region consistency loss is written as $\sum_{i=1}^{K} {L_c}_i$.  
Finally, the complete training objective is defined as:
\begin{equation}
    L_{\text{training}} = \alpha L_r + (1-\alpha) \sum_{i=1}^{K} {L_c}_i.
\end{equation}
As shown in Figure \ref{fig:fig_attention}, this training strategy enables identifiers to effectively attend to their corresponding local regions in the reference image. Ablation studies (Sec. \ref{sec: ablations}) further validate the effectiveness of our training objective.

\section{Experiments}
\label{sec:Experiments}
\subsection{Implementation Details}
\textbf{Dataset.} 
Consistent with prior reference guided generation works \cite{ouyang2025k,shah2024ziplora,liu2024unziplora}, we curate 500 evaluation images, ensuring direct comparability with existing methods. We first collect 60 diverse reference images from open-source repositories \cite{zhang2023prospect} and text-to-image models. We also extend single-object datasets used in previous works~\cite{sohn2023styledrop,ruiz2023dreambooth} to include multiple objects.  
For each reference image, we use 3–5 validation prompts and generate 3 samples per prompt, resulting in more than 500 evaluation images. At each iteration, we randomly select 200 images to report evaluation metrics.  

\noindent \textbf{Experimental Setup.} 
We use Stable Diffusion 3.5 and Flux as our DiT-based generation models. We adopt GPT-4o as the MLLM \cite{nie2024mmrel} for triplet extraction and selection in SCM. We additionally evaluate with an open-source MLLM (InternVL, sec \ref{sec: ablations}) to verify robustness across extraction models. Following the block heuristics in~\cite{huggingface_diffusers_dreambooth_readme}, we train LoRA blocks for layers 12–24 and 30–37, while keeping the base model weights and text encoder frozen.  
The LoRA weights are trained with a rank of 32, $\alpha = 0.5$, using the Adam optimizer (learning rate = $5\times10^{-5}$) for 750 steps with a batch size of 1. More details on the runtime and hardware specifications are provided in the supplementary material.
\subsection{Qualitative Comparisons}
\label{subsec:qual_comp}
\begin{figure*}[t]
    \centering
    \includegraphics[width=\linewidth]{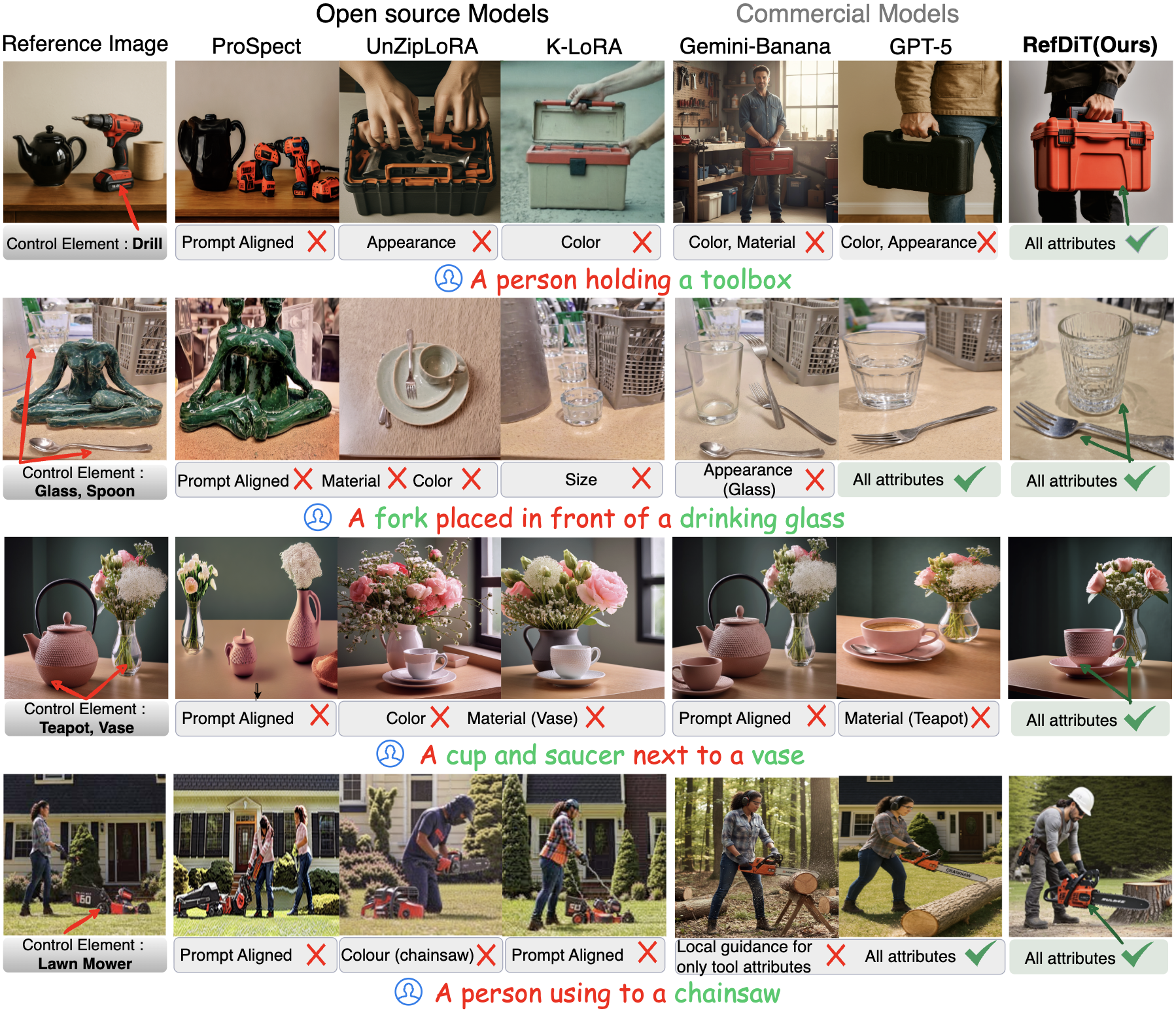}
    \caption{\textbf{Qualitative comparison} on complex multi-object scenes. 
Open-source methods rely on global conditioning and often transfer attributes from incorrect regions or partially ignore the prompt. 
Commercial models generate realistic images but may overfit to reference objects not specified in the prompt. 
In contrast, RefDiT selectively transfers attributes from the local elements while preserving prompt consistency.} 
    \vspace{-0.6cm}
    \label{fig:comparison_1}
\end{figure*}  
Figure~\ref{fig:comparison_1} shows the results of our method on real-world scenes containing multiple objects, where the validation prompts include objects that match local elements in the reference image. Row 1 in Figure~\ref{fig:comparison_1} shows that, given a reference containing a \textit{\textcolor{green!50!black}{drill}} and an inference prompt ``\textit{\textcolor{red}{A person holding a} \textcolor{green!50!black}{toolbox}} '', RefDiT accurately guides the generation using the drill’s color, material, and appearance. In contrast, UnZipLoRA fails to replicate the red color, while Gemini-Banana and GPT-5 generate single-colored outputs, miss the two color design of the drill, and lack material and appearance consistency. As shown in Rows 2 and 3, our method achieves performance competitive with commercial models. In Row 3, Gemini-Banana fails to align with the input prompt, as it generates the ``teapot'' not present in the prompt. GPT-5 better preserves the color alignment, but our method maintains superior material details and pattern quality of the teapot. Row 4 shows that existing methods tend to overfit to the original scene by generating nearly identical copies of the same person. In contrast, our method produces a similar tool while generating a different person, demonstrating better local guidance without identity overfitting.

\begin{figure}[h]
    \centering
    \begin{subfigure}[t]{0.45\textwidth}
        \centering
        \includegraphics[
            height=0.26\textheight,
            keepaspectratio
        ]{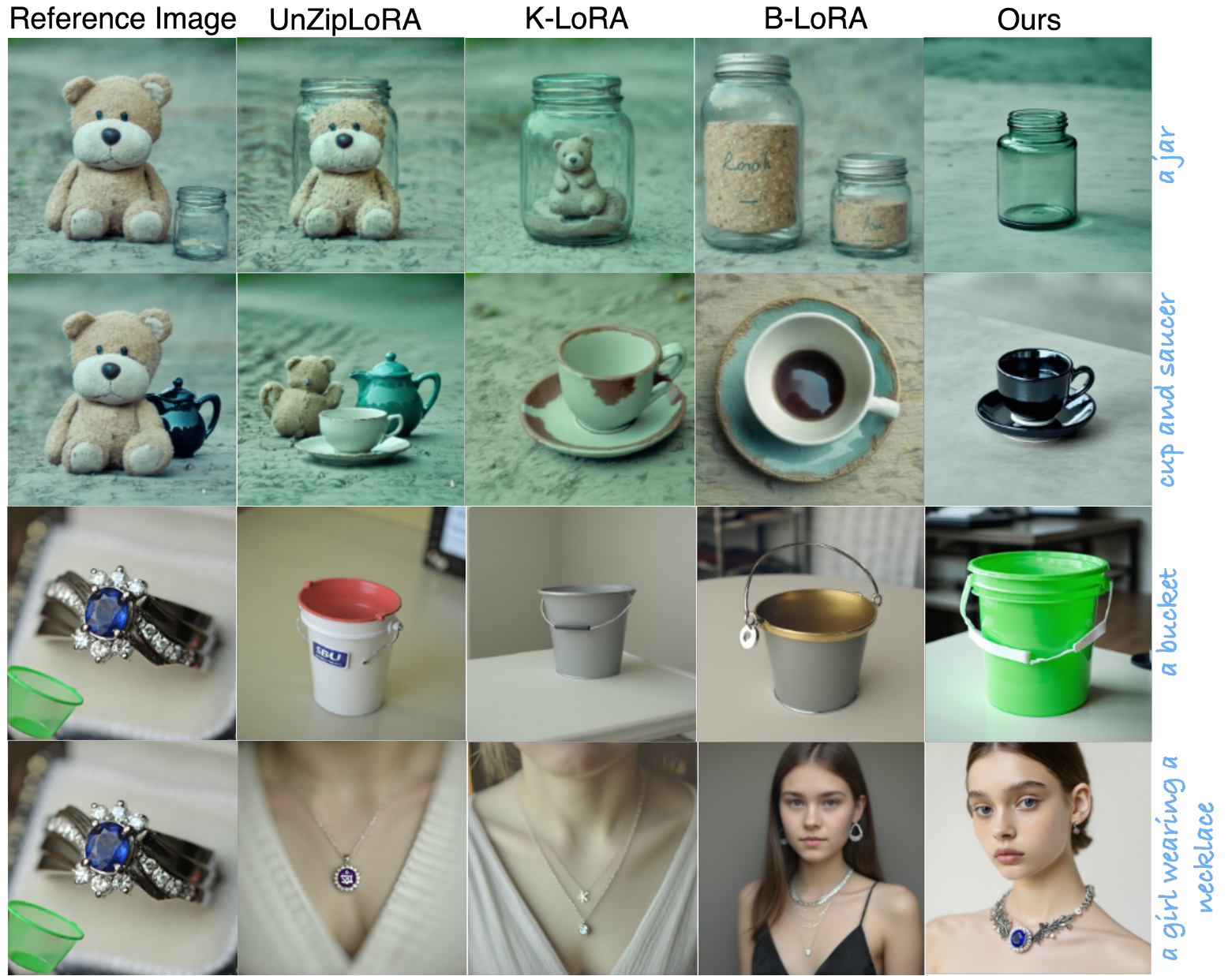}
        \caption{}
        \label{fig:comarison_2}
    \end{subfigure}
    \hfill
    \begin{subfigure}[t]{0.46\textwidth}
        \centering
        \includegraphics[
            height=0.33\textheight,
            width=\linewidth,
            keepaspectratio
        ]{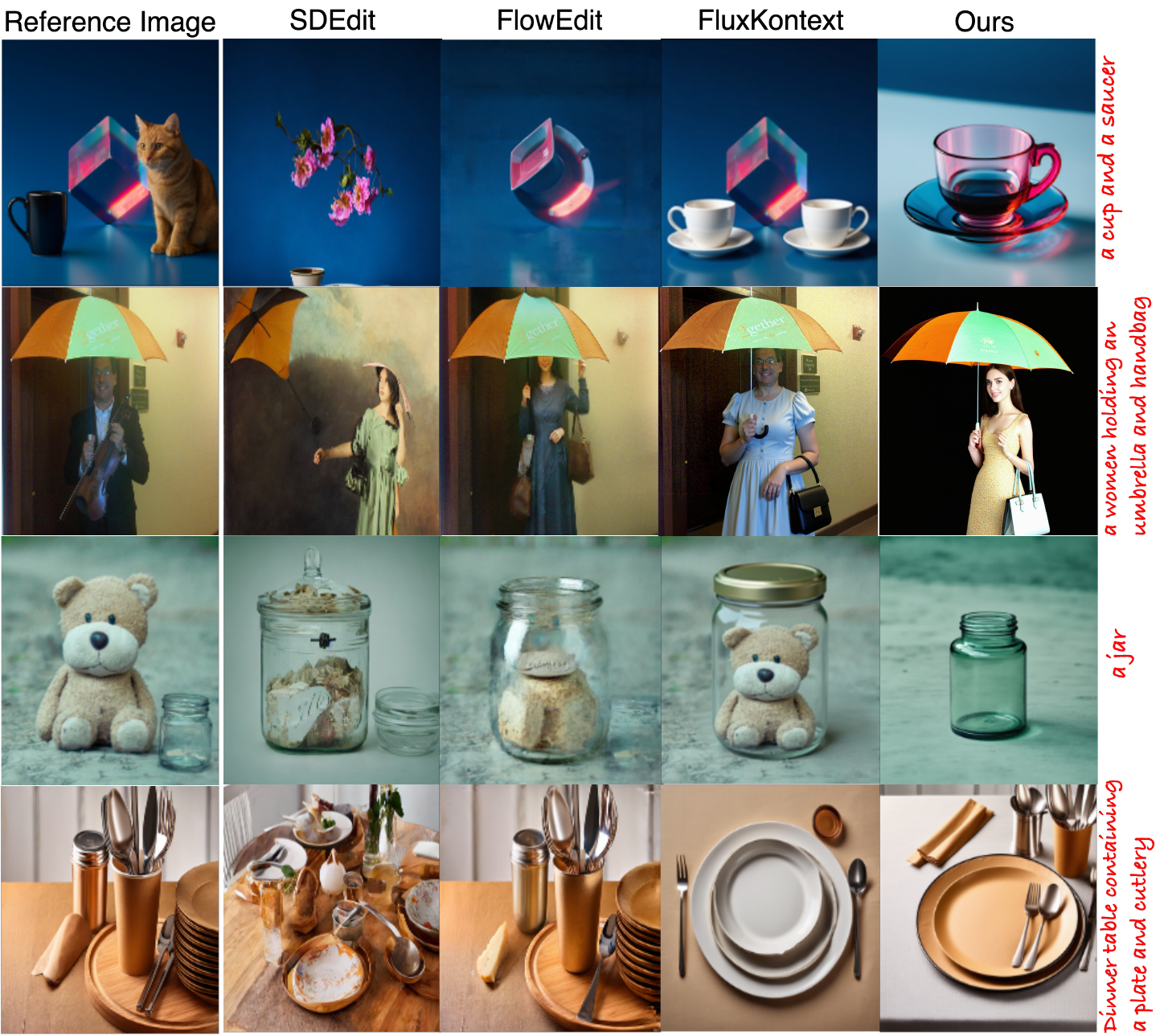}
        \caption{}
        \label{fig:comarison_3}
    \end{subfigure}
    \caption{(a) Visual illustration of the limited localization capabilities of existing personalization methods when single-object datasets are extended to include multiple objects. (b) Comparison of RefDiT with editing models, highlighting their limited adaptability in preserving local attributes during novel image generation. The corresponding inference prompts are shown at the end of each row.}
    \vspace{-30pt}
    \label{fig:comparison_combined}
\end{figure}

Existing methods perform well for single-object cases, but struggle with localizing reference guidance. To evaluate this, we extend single-object datasets from prior work by introducing additional objects with distinct visual attributes using commercial inpainting models \cite{AdobePhotoshop}, creating more complex evaluation samples.  
Figure~\ref{fig:comarison_2} shows results for such examples. For instance, in rows 1 and 2, we add a container and a teapot to a popular toy reference image. RefDiT generates outputs consistent with local regions of the reference.
Figure~\ref{fig:comarison_3} compares our results with various image editing methods, supporting our claim that such methods are not adaptable to our task. As shown in row 1 of Figure \ref{fig:comarison_3}, editing methods fail when the user-provided control requires guiding the generation process using the attributes of the cube. In this case, all editing methods were unable to generate a cup and saucer consistent with the cube’s attributes, demonstrating their inability to preserve relevant local attributes when generating novel images. Additional qualitative comparisons are provided in the supplementary.
\subsection{Comparative Analysis}
\label{subsec:comp_analysis}
We formulate our task as reference-guided generation, where attribute-level details of local regions in a single reference image guide the generation process in complex scenes containing multiple objects.  
To the best of our knowledge, there is no existing DiT-based approach that addresses reference guidance in this manner. The most relevant baselines are LoRA-based personalization models, which learn implicit content and style concepts from a reference image to generate contextually guided outputs.
\begin{table}[h]
\centering
\caption{Comparison of RefDiT with existing methods. \textbf{Bold} indicates best overall performance and \underline{underline} indicates best open-source baseline performance.}
\begin{tabular}{lcccc}
\toprule
 & CLIP-I$\uparrow$  & CLIP-T$\uparrow$  & Attr-Match$\uparrow$  & Attr-SIM$\uparrow$  \\
\midrule
DLoRA \cite{ruiz2023dreambooth}     & 0.71 & 0.29 & 0.48 & 0.41 \\
B-LoRA \cite{frenkel2024implicit}     & 0.70 & 0.29 & 0.42 & 0.44 \\
K-LoRA \cite{ouyang2025k}     & 0.77 & 0.27 & 0.58 & 0.50 \\
UnZipLoRA \cite{liu2024unziplora}  & 0.73 & \underline{0.30} & 0.52 & 0.46 \\
ProSpect \cite{zhang2023prospect}  & 0.68 & 0.21 & 0.28 & 0.33 \\
SDEdit \cite{meng2021sdedit}     & 0.75 & 0.22 & 0.33 & 0.35 \\
FlowEdit \cite{kulikov2024flowedit}   & 0.73 & 0.24 & 0.38 & 0.38 \\
FluxKontext          & \underline{0.79} & 0.28 & \underline{0.60} & \underline{0.52} \\
\midrule
\multicolumn{5}{l}{\textit{\textcolor{gray}{Commercial Models}}} \\
\textcolor{gray}{Gemini-Banana}  & \textcolor{gray}{0.78} & \textcolor{gray}{0.29} & \textcolor{gray}{0.74} & \textcolor{gray}{0.65} \\
\textcolor{gray}{GPT-5}          & \textcolor{gray}{0.82} & \textcolor{gray}{0.31} & \textcolor{gray}{0.84} & \textcolor{gray}{0.70} \\
\midrule
\rowcolor{green!15}
Ours (SD 3.5 DiT) & 0.80 & \textbf{0.32} &
\textbf{0.88} &
0.70 \\

\rowcolor{green!15}
Ours (Flux DiT) & \textbf{0.83} & 0.31 &
0.86 &
\textbf{0.74} \\
\bottomrule
\end{tabular}
 \vspace{-20pt}
\label{tab:scores}
\end{table}

These LoRA-based methods perform well in capturing the overall style of a reference image and can implicitly learn certain attributes such as color and material.  
We compare our method with DreamBooth-LoRA (DLoRA) \cite{ruiz2023dreambooth} using the same DiT architecture.  
For B-LoRA \cite{frenkel2024implicit}, and K-LoRA \cite{ouyang2025k}, we discard the content LoRA blocks since we aim to generate new content, and use only the style blocks for guidance.  
Similarly, in UnZipLoRA \cite{liu2024unziplora}, we decompose the reference, and discard content-related details.
We also evaluate image editing methods to show that they cannot be directly adapted for our task. We include DiT-based editing models such as FluxKontext \cite{labs2025flux} and FlowEdit \cite{kulikov2024flowedit}, as baselines.

Following~\cite{liu2024unziplora,ouyang2025k,zhong2024multi}, we report CLIP image similarity (CLIP-I) and text similarity (CLIP-T) scores for the generated images.  
While these scores measure overall alignment with the reference image and text prompt, they do not capture attribute-level correspondence.  
To address this, we introduce the Attr-Match score, which measures the attribute-level match between relevant local regions in the reference and generated images. 
We compute Attr-Match score using MLLM to extract attribute details for five key attributes, color, material, shape, size, and global appearance.
Since language models can produce varied textual descriptions for the same attribute, we restrict their output by defining a vocabulary of attribute values from the test set. The model must select from this vocabulary or return an “out-of-vocabulary” value. 
Attr-Match is defined as the ratio of matched attributes for corresponding objects between the reference and generated images. We manually validate the Attr-Match score on a subset of test combinations to ensure reliability. We use LLaMA3.2-11B for Attr-Match computation, distinct from GPT-4o used in SCM extraction as prior work on LLM-as-a-judge \cite{li2025preference} indicates that this combination exhibits minimal preference leakage.
\begin{wrapfigure}{l}{0.45\textwidth}  
\vspace{-20pt}
    \centering
    \includegraphics[width=\linewidth]{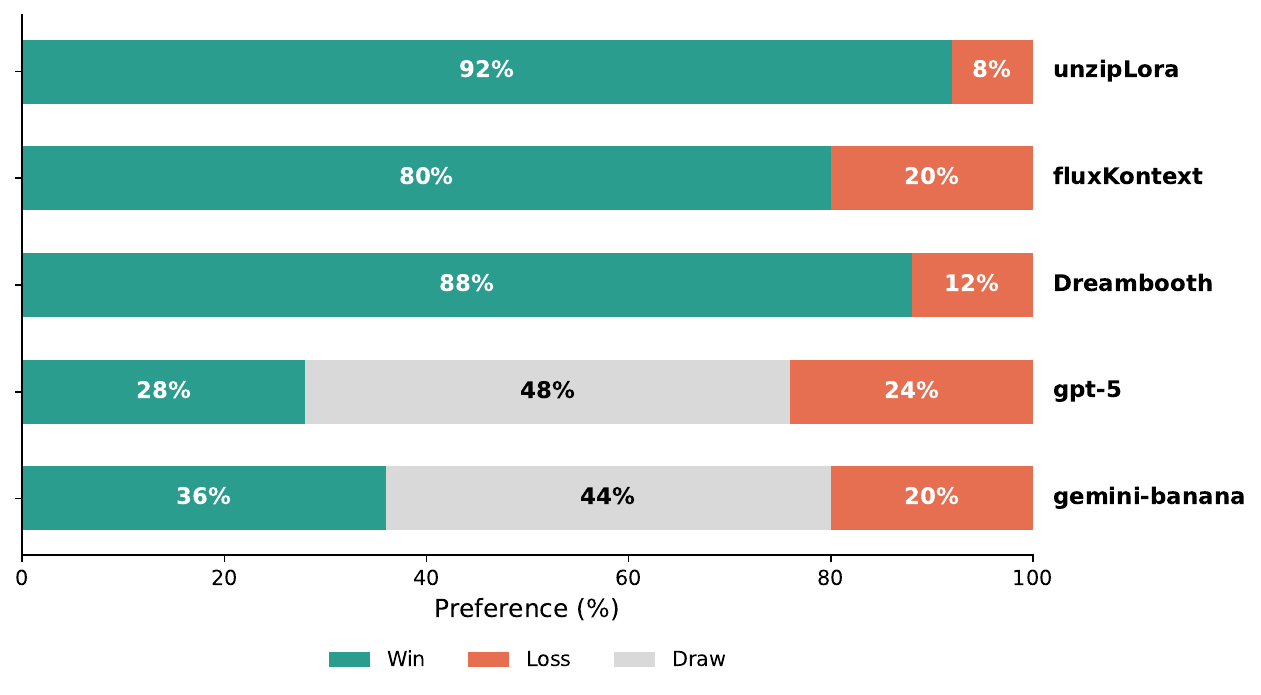}
    \caption{\textbf{Human User Study:} Green portion of each bar represents the percentage of responses preferring our outputs over the corresponding baseline.}
    \vspace{-20pt}
    \label{fig:fig_graph}
\end{wrapfigure}
Similar to CLIP-T, we also report the Attr-SIM score. Here, we query a language model to describe corresponding local regions in both the reference and generated images using attribute-based captions and compute their text similarity. Attr-SIM captures cases where semantically similar attributes are described using different words, complementing Attr-Match. Attr-SIM uses free-form descriptions, not constrained on vocabulary.  

\textbf{Results.} 
Table~\ref{tab:scores} presents the quantitative comparison of our method with relevant baselines.  
Based on CLIP-I and CLIP-T scores, our method generates images consistent with both the reference image and the inference prompt.  
We observe significant improvements in Attr-Match and Attr-SIM scores compared to open-source baselines, demonstrating effective attribute-based local guidance by RefDiT.  
Our results are also competitive with large-scale commercial models such as OpenAI’s GPT-5 and Google’s Gemini-Banana.

\begin{wrapfigure}{l}{0.4\textwidth} 
    \vspace{-20pt}
    \centering
    \includegraphics[width=\linewidth]{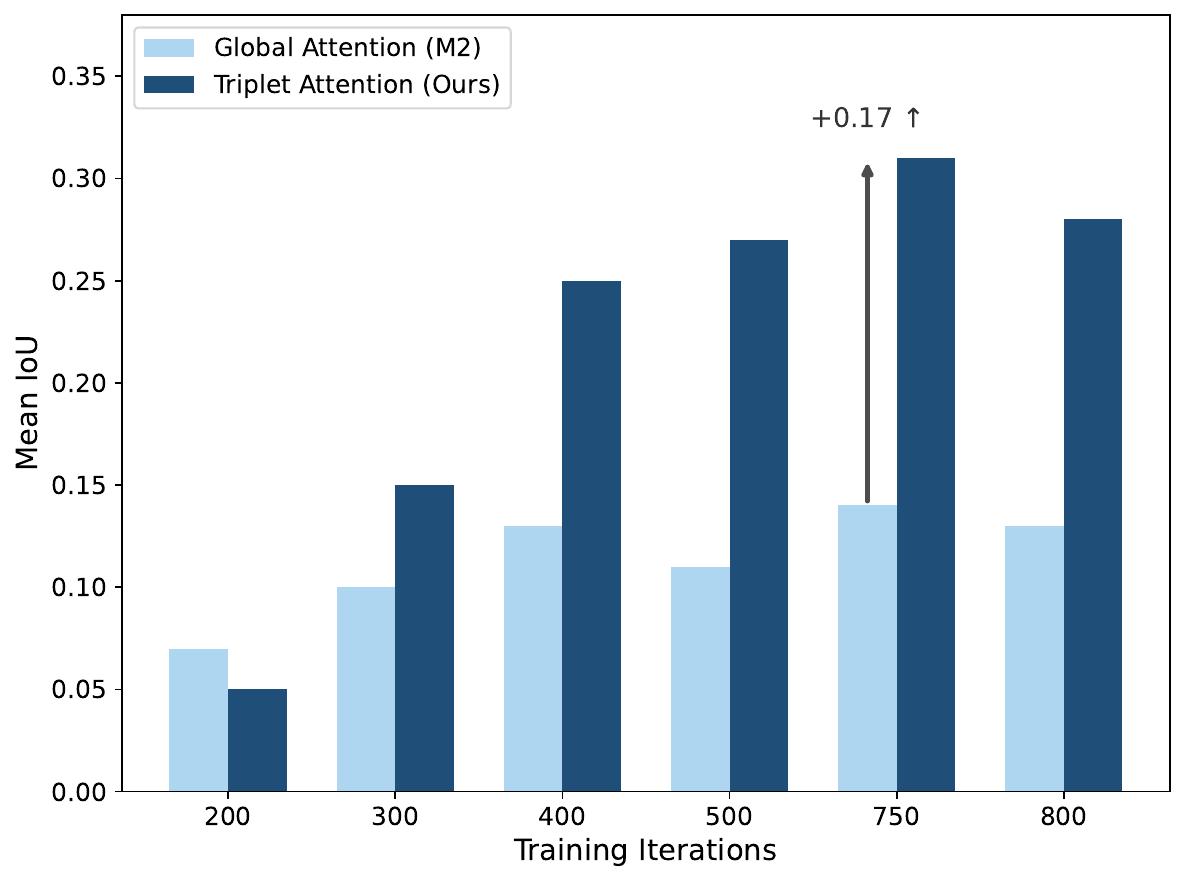}
    \caption{Mean IoU of identifier-token attention maps in M2 versus triplet-conditioned RefDiT across training iterations.}
    \vspace{-20pt}
    \label{fig:fig_miou}
\end{wrapfigure}

\textbf{Human User Study.} To further evaluate our method and assess human perception of the generated results, we conducted a user study.  
We prepared two types of questionnaires, one asking participants to select outputs that best align with the relevant region in the reference image, and another asking them to verify whether individual attributes match between the local regions of the reference and generated images. A total of 45 participants submitted their responses, each answering 10 questions, resulting in 450 responses overall.   
Figure~\ref{fig:fig_graph} shows the results of our user study. Participants preferred RefDiT over UnZipLoRA (92\%), DreamBooth (88\%), and FluxKontext (80\%). When compared to GPT-5, 28\% preferred RefDiT while 48\% rated the outputs as comparable in quality. We provide additional analyses, results, and failure cases of our method in the supplementary material.
\vspace{-15pt}
\subsection{Ablation Analysis}
\label{sec: ablations}
\begin{wraptable}{r}{0.52\textwidth}
    \vspace{-35pt}
    \caption{Ablation results showing the contribution of each component. \textcolor{green!60!black}{\checkmark} indicates presence of the component, \textcolor{red}{$\times$} indicates absence.}
    \centering
    \setlength{\tabcolsep}{4pt}
    \renewcommand{\arraystretch}{1.05}
    \resizebox{\linewidth}{!}{
    \begin{tabular}{lccc|ccc}
    \toprule
     & SCM & Triplet Cond. & $\alpha$ & CLIP-I$\uparrow$ & CLIP-T$\uparrow$ & Attr-SIM$\uparrow$ \\
    \midrule
    M1 & \textcolor{red}{$\times$} & \textcolor{red}{$\times$} & \textcolor{red}{$\times$} & 0.71 & 0.29 & 0.43 \\
    M2 & \textcolor{green!60!black}{\checkmark} & \textcolor{red}{$\times$} & \textcolor{red}{$\times$} & 0.77 & 0.27 & 0.57 \\
    M3 & \textcolor{green!60!black}{\checkmark} & \textcolor{green!60!black}{\checkmark} & 0.7 & 0.79 & 0.32 & 0.68 \\
    M4 & \textcolor{green!60!black}{\checkmark} & \textcolor{green!60!black}{\checkmark} & 0.3 & 0.80 & 0.30 & 0.66 \\
    M5 & \textcolor{green!60!black}{\checkmark} (InternVL) & \textcolor{green!60!black}{\checkmark} & 0.3 & 0.73 & 0.30 & 0.63 \\
    \rowcolor{green!15} Ours & \textcolor{green!60!black}{\checkmark} (GPT-4o) & \textcolor{green!60!black}{\checkmark} & 0.5 & 0.80 & 0.32 & 0.70 \\
    \bottomrule
    \end{tabular}}
    \vspace{-20pt}
    \label{tab:ablation}
\end{wraptable}

In this section, we provide additional analysis to demonstrate the effectiveness of each component in RefDiT. Table~\ref{tab:ablation} reports the quantitative results of these ablations.

\noindent \textbf{Impact of SCM.} In M1, we adopt the DreamBooth setup, which uses a single identifier token and a reconstruction loss. In M2, we integrate our SCM module into M1 without including the triplet conditioning. The improvement from M1 to M2 demonstrates that using a structurally attribute-aware conditioning signal enhances guidance.

\noindent \textbf{Impact of Triplet Conditioning.} The improvement from M2 to our final model indicates that the triplet conditioned training provides more effective local guidance.
Furthermore, Figure \ref{fig:fig_miou} shows that the mean Intersection over Union (mIoU) of cross-attention masks corresponding to identifier tokens improves by 0.17 compared to M2, which uses only global conditioning. 
The IoU scores are computed with respect to the original object masks obtained using SAM \cite{kirillov2023segany}.

\noindent \textbf{Choosing $\alpha$.}
 In M3, we set $\alpha = 0.7$, and in M4, $\alpha = 0.3$. M3 and M4 show that setting $\alpha$ higher or lower than 0.5 leads to reduced performance. 

\noindent \textbf{Robustness of SCM Extraction.}
 In M5, to evaluate SCM’s dependence on the MLLM, we replace GPT-4o with the open-source InternVL-3.5 \cite{wang2025internvl3_5} to extract attribute-level triplets.
 
 \begin{wraptable}{r}{0.4\textwidth}
    \vspace{-12pt}
    \caption{User study preferences comparing different variants.}
    \centering
    \resizebox{\linewidth}{!}{
    \begin{tabular}{lcc}
    \toprule
    \multicolumn{3}{c}{\textbf{\% Preference for:}} \\
    \midrule
    \textbf{M2 over M1} & \textbf{Ours over M2} & \textbf{Ours over M4} \\
    \midrule
    72.8\% & 68.7\% & 58.4\% \\
    \bottomrule
    \end{tabular}}
    \label{tab:ablation_user}
    \vspace{-20pt}
\end{wraptable}
 As shown in Table~\ref{tab:ablation} (M5), InternVL yields competitive results (Attr-Match 0.80, Attr-SIM 0.63), validating that SCM’s triplet-based conditioning is effective even with open-source MLLMs.
 
 We also conducted a user study for our ablations. Table~\ref{tab:ablation_user} shows that users preferred the outputs of our complete method over variants with modified configurations.
\vspace{-10pt}
\section{Discussion}
\noindent \textbf{SCM Validation}: To assess the reliability of attribute-based triplet extraction, we manually validate a randomly selected subset of 50 images from our test set. Averaged over five key attributes, SCM achieves 90.5\% attribute accuracy (96.2\% for color and 89.7\% for material) and 94\% accuracy for object classification. In supplementary, we provide a detailed analysis of the limitations and sensitivity to failure cases of using an MLLM for SCM, along with simulating SCM extraction failures. Even under simulated SCM extraction errors, performance remains above open-source baselines, due to its effective mapping of identifier tokens to local visual regions. Nevertheless, errors in object detection, attribute labeling, or relation parsing can propagate into conditioning signals, particularly in scenes with rare objects, ambiguous attributes, or heavy occlusion.

\noindent \textbf{Evaluation limitations}: Evaluating local attribute correspondence remains challenging. While RefDiT achieves strong performance on global alignment metrics such as CLIP-I and CLIP-T, these measures are insufficient for assessing region-level guidance. To address this, we introduce Attr-Match and Attr-SIM, which quantify attribute-level consistency between corresponding local regions. Since these metrics rely on language-model-based extraction and constrained vocabularies, they may introduce bias or overlook subtle perceptual nuances. To reduce circular evaluation effects, we use an MLLM (LLaMA), distinct from GPT-4o used in SCM extraction. Human studies further complement automatic metrics; however, scalable and fully objective evaluation of fine-grained local guidance remains an open research problem.
\vspace{-10pt}
\section{Conclusion}
We propose RefDiT, a method that enables effective local, attribute-level guidance for reference-constrained image generation. Through an extensive evaluation of existing reference-guided image generation models, we identify a key limitation when users attempt to generate new images guided by multiple local attributes in the reference image. To address this, we introduce a local attribute-guided, LoRA-based training strategy for DiT-based image generation models. Our experiments demonstrate that RefDiT improves reference guidance across a diverse set of references and provides users with greater control over the generation process. We validate the effectiveness of our method using multiple evaluation metrics and support our findings with a user study that highlights the perceptual quality of the generated outputs.

\clearpage
%
%
\bibliographystyle{splncs04}
\bibliography{main}
\end{document}